\documentclass[conference]{IEEEtran}
\IEEEoverridecommandlockouts
\usepackage{cite}
\usepackage{amsmath,amssymb,amsfonts}
\usepackage{algorithmic}
\usepackage{graphicx}
\usepackage{textcomp}

\usepackage{multirow}
\usepackage{tabularx,booktabs}
\newcolumntype{C}{>{\centering\arraybackslash}X} 
\usepackage[dvipsnames]{xcolor}
\usepackage{longtable}
\usepackage{makecell, cellspace, caption}
\usepackage{array}
\usepackage{subcaption}
\usepackage{cuted}
\usepackage{multirow}
\usepackage[utf8]{inputenc}
\usepackage{tabularx}
\usepackage{array}
\usepackage{amssymb}
\usepackage{makecell}
\usepackage[margin=2cm]{geometry}
\usepackage{adjustbox}
\usepackage{multirow}
\usepackage{tabulary,booktabs}
\usepackage{ragged2e}
\usepackage{amsmath}
\usepackage{booktabs}
\usepackage{array}
\newcolumntype{L}{>{$}l<{$}}
\newcolumntype{C}{>{$}c<{$}}
\newcolumntype{R}{>{$}r<{$}}

\usepackage{amssymb}
\usepackage{pifont}
\usepackage{graphicx}
\usepackage[caption=false]{subfig}
\usepackage{url}
\usepackage{flushend}

\def\BibTeX{{\rm B\kern-.05em{\sc i\kern-.025em b}\kern-.08em
    T\kern-.1667em\lower.7ex\hbox{E}\kern-.125emX}}
\begin{document}

\title{INN-PAR: Invertible Neural Network for PPG to ABP Reconstruction}

\author{\IEEEauthorblockN{Soumitra Kundu}
\IEEEauthorblockA{\textit{Rekhi Centre of Excellence for the Science of Happiness} \\
\textit{IIT Kharagpur, India}\\
soumitra2012.kbc@gmail.com}
\and
\IEEEauthorblockN{Gargi Panda}
\IEEEauthorblockA{\textit{Department of EE} \\
\textit{IIT Kharagpur, India}\\
pandagargi@gmail.com}
\and
\IEEEauthorblockN{Saumik Bhattacharya}
\IEEEauthorblockA{\textit{Department of E\&ECE} \\
\textit{IIT Kharagpur, India}\\
saumik@ece.iitkgp.ac.in}
\and
\IEEEauthorblockN{}
\IEEEauthorblockA{ \\
\\
}
\and
\IEEEauthorblockN{Aurobinda Routray}
\IEEEauthorblockA{\textit{Department of EE} \\
\textit{IIT Kharagpur, India}\\
aroutray@ee.iitkgp.ac.in}
\and
\IEEEauthorblockN{Rajlakshmi Guha}
\IEEEauthorblockA{\textit{Rekhi Centre of Excellence for the Science of Happiness} \\
\textit{IIT Kharagpur, India}\\
rajg@cet.iitkgp.ac.in}
}

\maketitle

\begin{abstract}
Non-invasive and continuous blood pressure (BP) monitoring is essential for the early prevention of many cardiovascular diseases. Estimating arterial blood pressure (ABP) from photoplethysmography (PPG) has emerged as a promising solution. However, existing deep learning approaches for PPG-to-ABP reconstruction (PAR) encounter certain information loss, impacting the precision of the reconstructed signal. To overcome this limitation, we introduce an invertible neural network for PPG to ABP reconstruction (INN-PAR), which employs a series of invertible blocks to jointly learn the mapping between PPG and its gradient with the ABP signal and its gradient. INN-PAR efficiently captures both forward and inverse mappings simultaneously, thereby preventing information loss. By integrating signal gradients into the learning process, INN-PAR enhances the network’s ability to capture essential high-frequency details, leading to more accurate signal reconstruction. Moreover, we propose a multi-scale convolution module (MSCM) within the invertible block, enabling the model to learn features across multiple scales effectively. We have experimented on two benchmark datasets, which show that INN-PAR significantly outperforms the state-of-the-art methods in both waveform reconstruction and BP measurement accuracy.
Codes can be found at: \url{https://github.com/soumitra1992/INNPAR-PPG2ABP}.
\end{abstract}

\begin{figure*}[hbt!] 
    \centering
  {\includegraphics[width=0.9\linewidth]{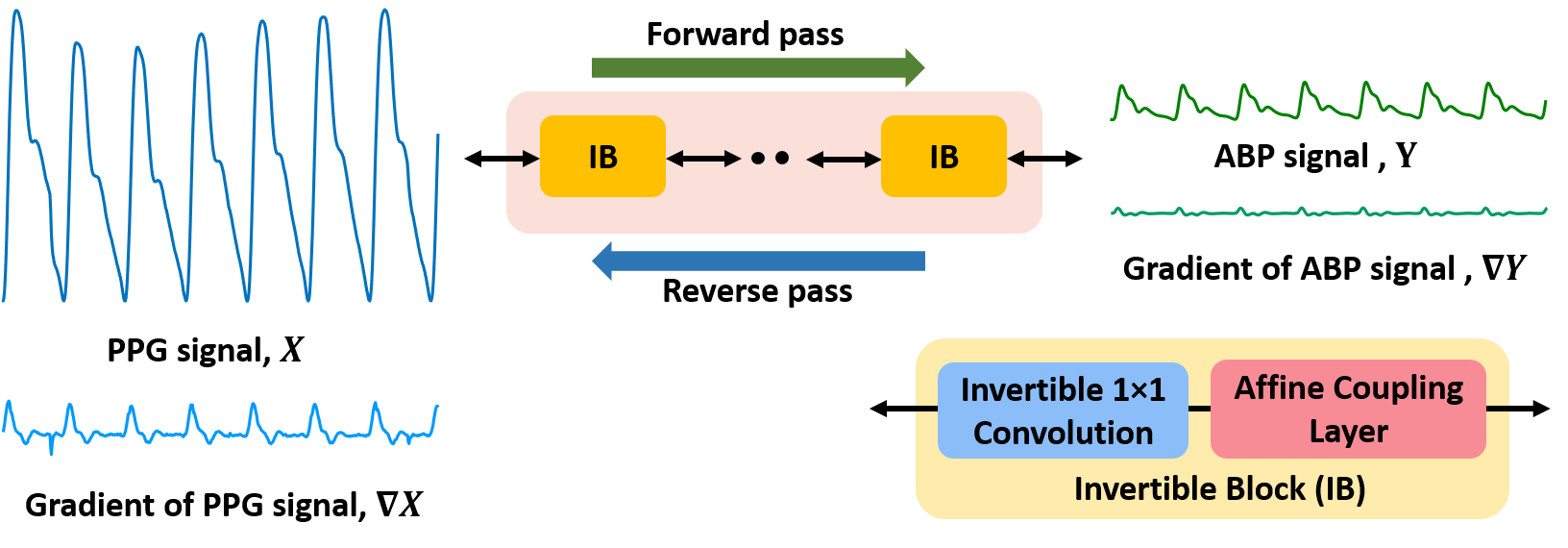}}
  \caption{INN-PAR architecture and Invertible Block (IB) structure.}
  \label{fig1} 
\end{figure*}
\section{Introduction}
\label{sec:intro}

Continuous monitoring of blood pressure (BP) is essential for the early detection and management of several cardiovascular diseases. In BP measurement, systolic blood pressure (SBP), the higher of the two values, signifies the peak pressure in the arteries during the heart's contraction, while diastolic blood pressure (DBP) indicates the minimum pressure during the heart's resting phase between beats. Both SBP and DBP values are expressed in millimeters of mercury (mmHg). Continuous monitoring aids in adjusting medications and lifestyle modifications for improved BP management.

Traditional methods of measuring BP can be divided into invasive and non-invasive techniques. Invasive techniques, which involve arterial cannulation, pose serious risks, while non-invasive methods like cuff-based sphygmomanometers offer a safer, more convenient alternative. However, these cuff-based devices are often bulky and impractical for continuous monitoring due to the need for repeated cuff inflations and deflations.

To address these limitations, photoplethysmography (PPG) has emerged as a promising technique for measuring BP in a continuous and non-invasive manner. Wearable devices like smartwatches, fitness bands, and smart rings have significantly simplified the collection of PPG signals. Aditionally, recent advancements in deep learning have enabled the continuous extraction of arterial blood pressure (ABP) waveforms from PPG signals. BP is estimated by analyzing these ABP waveforms to determine SBP and DBP by identifying the peak and crest in each cardiac cycle \cite{nature}.

Several studies have explored the reconstruction of ABP waveforms from PPG signals using deep learning techniques, such as convolutional neural networks (CNNs) \cite{sensors,unet1, vnet}, variational autoencoders (VAEs) \cite{vae}, and CycleGANs \cite{gan}. The U-Net architecture \cite{unet}, with its contracting and expansive paths connected by skip connections, has been widely adopted for ABP estimation from PPG signals \cite{sensors,unet1}. U-Net effectively translates one signal to another by combining contextual information and high-resolution features to reconstruct ABP waveforms. 
Hill \textit{et al.} \cite{vnet} proposed another method utilizing a modified V-Net architecture for predicting ABP waveforms.
Generally, these PPG-to-ABP reconstruction (PAR) methods \cite{sensors,vnet,unet1} involve two steps: feature extraction and signal reconstruction. Despite their effectiveness, these methods often experience significant information loss during the feature extraction stage due to the non-injective mapping learned by the network. Additionally, these methods do not account for the signal gradient, which contains critical structural information.

To address the challenge of information loss in deep neural networks, invertible neural networks (INN) have been proposed \cite{nice,irevnet}. INNs retain all information from the input signal by learning reversible bijective transformations, ensuring a complete and accurate mapping between the input signal $x$ and output signal $y$. 
Through a series of invertible and bijective functions $\{f_k\}_{k=0}^N$, $x$ and $y$ can be related as,

\begin{align}
\begin{split}
y & = f_0\circ f_1\circ f_2\circ \dots \circ f_N (x) \\
x & = f_0^{-1}\circ f_1^{-1}\circ f_2^{-1}\circ \dots \circ f_N^{-1} (y)
\end{split}
\end{align}

Though INNs have primarily been used for generative tasks in image \cite{nice,realnvp,glow} and signal processing \cite{waveglow,speech,aeflow}, several studies have also applied INNs for image reconstruction tasks \cite{rescaling,icassp7,colorflow,iisp}. In a similar manner, INNs may be advantageous for signal-to-signal reconstruction tasks, as they simultaneously learn both the forward and inverse mappings, which helps prevent information loss. Additionally, incorporating the signal gradient into the learning process, which focuses on the rate of change rather than just the signal values, enhances the ability to capture essential high-frequency details \cite{structure,sgnet}, thereby ensuring a more accurate reconstruction of the signal's shape.

In this paper, we propose a novel INN built using a stack of invertible blocks for the PAR task. While learning the mapping between PPG and ABP, INN-PAR simultaneously learns the mapping between their gradients, which improves the reconstruction capability of essential high-frequency details. Additionally, we introduce a multi-scale convolution module (MSCM) within the invertible blocks to capture multi-scale features effectively. During training the network, we constrain both the estimated ABP and its gradient to match the ground truth signals closely. Extensive experiments were conducted across two benchmark datasets, demonstrating that INN-PAR outperforms the state-of-the-art (SOTA) methods in both PPG-to-ABP waveform reconstruction as well as SBP and DBP measurement accuracy.
 
The rest of this paper is organized in the following manner. The architecture of INN-PAR, the structure of the invertible block (IB), and the loss function details are described in Section \ref{sec:method}. 
Experiments on two benchmark datasets show the effectiveness of our proposed method, which is demonstrated in Section \ref{sec:experiments}. Finally, we conclude the paper in Section \ref{sec:conclusion}. 
\begin{figure}[hbt!] 
    \centering
  {\includegraphics[width=0.85\linewidth]{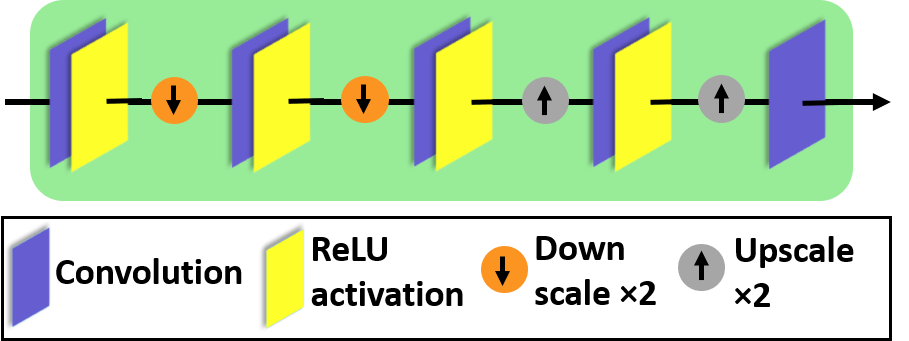}}
  \caption{Multi-Scale Convolution Module (MSCM).}
  \label{fig2} 
\end{figure}
\section{Proposed Method}
\label{sec:method}
\begin{figure*}[hbt!] 
   \includegraphics[width=\linewidth]{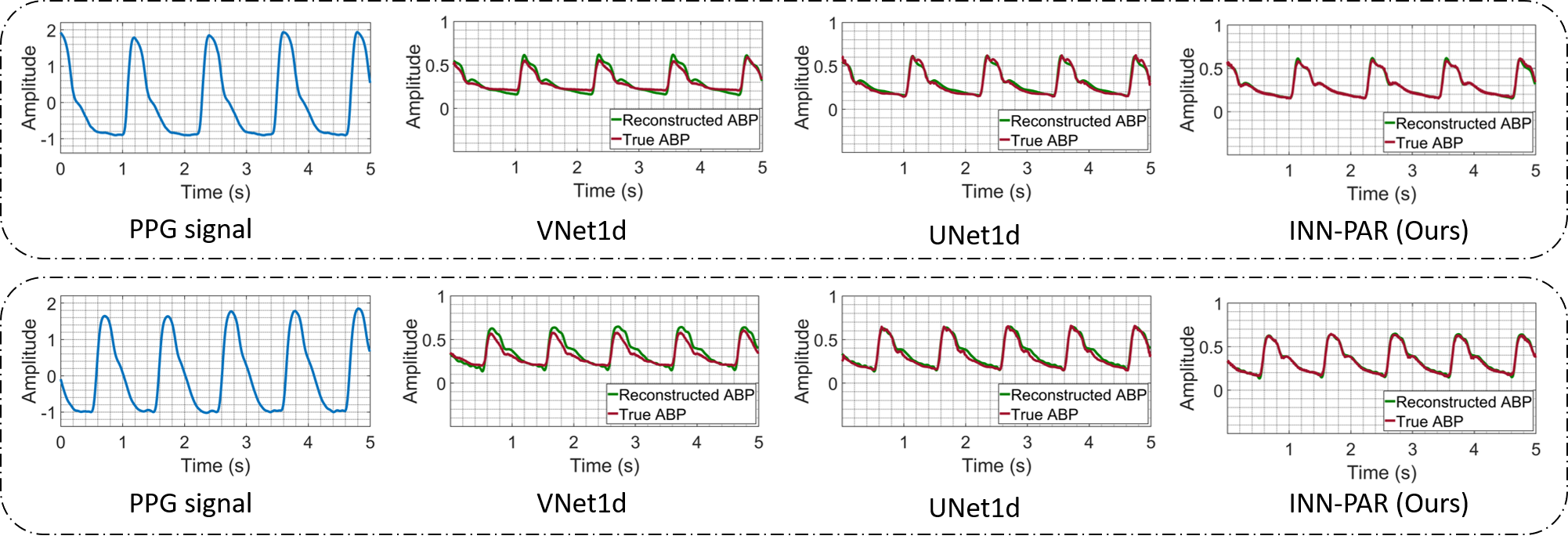}
  \caption{Visual comparison with SOTA methods. The first row shows a signal from the Sensors dataset \cite{sensors}, and the second row shows a signal from the BCG dataset \cite{bcg}. Signals are best viewed in $200\%$ zoom.}
  \label{fig3} 
\end{figure*}
\subsection{Overall architecture of INN-PAR}
Our proposed model INN-PAR aims to reconstruct an ABP signal $Y \in \mathbb{R}^{L \times 1}$ from a PPG signal $X \in \mathbb{R}^{L \times 1}$, where $L$ is the signal length. The ABP signal $Y$ is related to the PPG signal $X$ through a non-linear mapping: $f: Y=f(X)$. Hence, ABP to PPG signal reconstruction can be performed by the inverse mapping $f^{-1}: X=f^{-1}(Y)$. Our proposed method INN-PAR, utilizing an INN structure, simultaneously learns these two mappings. 

Following \cite{nice}, we design INN-PAR to have multi-channel input and output pairs. This allows the model to preserve all necessary information across the forward and inverse mappings. For the single-channel PPG and ABP signal pair, we incorporate their gradients as additional channels. Fig. \ref{fig1} illustrates the architecture of INN-PAR. While learning the mapping between $X$ and $Y$, we also learn the mapping between their gradients $\nabla X$ and $\nabla Y$, where $\nabla$ denotes the gradient operation. Since the gradient operation captures essential structural information and high-frequency details of the signal, learning gradient mappings helps INN-PAR more accurately reconstruct the high-frequency details of the ABP signal. Let $\mathcal{G}$ denote the INN-PAR, and we can write,
\begin{align}
\begin{split}
\text{forward pass,} \:\: \mathcal{G}& : (X,\nabla X) \longrightarrow (Y, \nabla Y) \\
\text{reverse pass,} \:\: \mathcal{G}^{-1}& : (Y, \nabla Y) \longrightarrow (X,\nabla X)
\end{split}
\end{align}

The invertibility of $\mathcal{G}$ is achieved by its invertible structure, containing a stack of invertible blocks (IBs). Each IB consists of an invertible $1\times1$ convolution followed by an affine coupling layer (ACL), described below.

\subsection{Affine Coupling Layer (ACL)}

ACL, the core component of IB, can operate in both forward and reverse modes. In the forward mode, ACL transforms an input pair $(X_1, X_2)$ to an output pair $(Y_1, Y_2)$ using the following equations \cite{rescaling}: 

\begin{align}
\begin{split}
Y_1 & = X_1+\mathcal{H}_1(X_2)\\
Y_2 & = X_2\odot exp(\mathcal{H}_2(Y_1))+\mathcal{H}_3(Y_1)
\end{split}
\end{align}

where $\odot$ denotes elementwise multiplication. $\mathcal{H}_1(\cdot)$, $\mathcal{H}_2(\cdot)$, and $\mathcal{H}_3(\cdot)$ are learnable modules, that need not be invertible \cite{rescaling}. In the reverse mode of ACL, $(Y_1, Y_2)$ is perfectly mapped back to $(X_1, X_2)$ as,

\begin{align}
\begin{split}
X_2 & =(Y_2-\mathcal{H}_3(Y_1))\oslash exp(\mathcal{H}_2(Y_1))\\
X_1 & = Y_1-\mathcal{H}_1(X_2)
\end{split}
\end{align}

where $\oslash$ denotes elementwise division. For $\mathcal{H}_1(\cdot)$, $\mathcal{H}_2(\cdot)$, and $\mathcal{H}_3(\cdot)$, we use multi-scale convolution module (MSCM) described below. 
\subsection{Multi-Scale Convolution Module (MSCM)}
Fig. \ref{fig2} shows the structure of MSCM, which processes the signal at multiple scales. At each scale, convolution layers are employed to extract the features. After each convolution operation, the ReLU activation function is used except for the last convolution layer.

For downscaling and upscaling of signals in MSCM, we do not use strided convolutions and deconvolution operations as they lead to loss of information. Instead, we design lossless squeeze and unsqueeze operations for the downscaling and upscaling steps. Like the Pixel Shuffle operation \cite{pixelshuffle}, the squeeze operation reduces the signal length while keeping the signal size the same by increasing the number of channels. Thus, the spatial information is preserved across the channel dimension. For the unsqueeze operation, we use the inverse of the squeeze operation, where the signal length is increased while keeping the signal size the same by reducing the number of channels.  
\subsection{Invertible $1\times1$ Convolution}
\label{inv}
Following Glow \cite{glow}, we use an invertible $1\times1$ convolution as the learnable permutation function to reverse the order of the channels before each ACL. The weights of this convolution are initialized to be orthonormal. This layer helps in mixing information across the channel dimension.
\subsection{Loss Function}
\label{loss}
During the training of INN-PAR, we constrain the reconstructed ABP signal $Y$ and its gradient $\nabla Y$ to match the ground truth (GT) ABP signal $Y_g$ and its gradient $\nabla Y_g$ using the $L_1$ loss function. The overall loss function is,
\begin{equation}\label{eqloss}
\mathcal{L}=||Y-Y_g||_1+\alpha ||\nabla Y-\nabla Y_g||_1
\end{equation}

where $\alpha$ is a tuning parameter. In our experiments, we set the value of $\alpha$ to $1$.
\begin{table*}[hbt]
\fontsize{9.5}{11.5}\selectfont
\centering
\caption{Performance comparison using two benchmark datasets with SOTA methods. FLOPs are measured for reconstructing an ABP signal with a length of $625$. The best values are highlighted. $\downarrow$ means a low value is desired. SBP and DBP values are in mmHg unit.}
\begin{tabular}{lcccccccccc}
\toprule
\multirow{3}{*}{Method} & \multirow{3}{*}{Params(K)} & \multirow{3}{*}{FLOPs(M)} & 
\multicolumn{4}{c}{Sensors \cite{sensors}} &
\multicolumn{4}{c}{BCG \cite{bcg}}\\
\cmidrule(lr){4-7}
\cmidrule(lr){8-11}
& & &
\multicolumn{2}{c}{Waveform}&
\multicolumn{1}{c}{SBP}& \multicolumn{1}{c}{DBP}&
\multicolumn{2}{c}{Waveform}&
\multicolumn{1}{c}{SBP}& \multicolumn{1}{c}{DBP}\\
\cmidrule(lr){4-5}
\cmidrule(lr){6-6}
\cmidrule(lr){7-7}
\cmidrule(lr){8-9}
\cmidrule(lr){10-10}
\cmidrule(lr){11-11}
& &
&\multicolumn{1}{c}{MAE$\downarrow$}&\multicolumn{1}{c}{NRMSE$\downarrow$}& \multicolumn{1}{c}{MAE$\downarrow$}& \multicolumn{1}{c}{MAE$\downarrow$}& \multicolumn{1}{c}{MAE$\downarrow$}& \multicolumn{1}{c}{NRMSE$\downarrow$}& \multicolumn{1}{c}{MAE$\downarrow$}& \multicolumn{1}{c}{MAE$\downarrow$}\\
\midrule
\text{VNet1d \cite{nature}} & 530   & 17.19 & 0.062 & 0.654 & 16.45 & 8.57 & 0.079 & 0.872 & 13.56 & 9.87 \\
\text{UNet1d \cite{nature}} & 141 & 43.76 & 0.059 & 0.527  & 15.34  & 7.29  & 0.080  & 0.725  & 12.74 & 8.02  \\
\multirow{1}{*}{INN-PAR (Ours)}  & 372 & 0.018 & \textbf{0.058}  & \textbf{0.506}  & \textbf{15.10} & \textbf{7.20}  & \textbf{0.075}  & \textbf{0.701}  & \textbf{11.96}  & \textbf{7.93} \\
\bottomrule
\end{tabular}
\label{tab:main1}
\end{table*}
\section{Experiments}
\label{sec:experiments}
\subsection{Experimental Setup}

For the experiments, we utilize two publicly available datasets: Sensors \cite{sensors} and BCG \cite{bcg}. We use the Sensors dataset for training and testing INN-PAR. Moreover, we also test the performance of INN-PAR on the BCG dataset to check the model's generalization ability, without any finetuning. We follow \cite{nature} for data pre-processing and splitting the dataset into training, validation, and test sets. We are not using any subject calibration and PPG scaling. Also, we do not allow subject information leakage among the training, validation, and test sets. The signals are segmented into $5s$ chunks without overlapping. For the Sensors dataset, we use $6658$ signal chunks for training, $2383$ for validation, and $2061$ for testing. For the BCG dataset, we use $671$ signal chunks for testing.
We train INN-PAR by minimizing the loss function in Eqn. \ref{eqloss} using Adam optimizer ($\beta_1=0.9,\: \beta_2=0.999$) for $500$ epochs with a batch size of $128$. The learning rate is kept constant at $1\times10^{-4}$. Using the Pytorch framework, we have conducted all the experiments in the NVIDIA A40 GPU. Mean absolute error (MAE) and normalized root mean square error (NRMSE) \cite{nrmse} metrics are used to compare the waveform reconstruction accuracy. For the accuracy of SBP and DBP measurement, we utilize the MAE metric.

For the implementation of INN-PAR, the number of invertible blocks is set to $4$. In MSCM, we process the signal at $3$ scales. At scale $1$, the number of convolution filters is $16$; at scale $2$, the number is $32$; and at scale $3$, the number is $64$. For all the $3$ scales, the kernel size for the convolution layer is set to $5\times1$. 
\subsection{Performance Comparison}
Since the majority of PPG to ABP reconstruction algorithms do not have open-source code, we have only been able to compare INN-PAR with two SOTA reconstruction methods. The algorithms used for comparison are VNet1d \cite{nature} and UNet1d \cite{nature}. To ensure a fair comparison, we utilized the models trained on the same dataset provided by the benchmark work \cite{nature}. 
Table \ref{tab:main1} exhibits a quantitative comparison of INN-PAR with the SOTA methods. In addition to the MAE and NRMSE metrics, the model parameters and FLOPs were also listed. We calculate the FLOPs under the setting of reconstructing an ABP signal of length $625$. A low value of model parameters, FLOPs, MAE, and NRMSE is desired. INN-PAR surpasses the SOTA methods in waveform reconstruction and SBP and DBP measurement performance. Though UNet1d has lower model parameters than INN-PAR, our method has significantly lower FLOPs or computational complexity. 

Figure \ref{fig3} shows the visual comparison of the PAR task for a signal from the Sensors and BCG datasets. VNet1d does not reconstruct the ABP signal well. UNet1d and INN-PAR show better signal reconstruction capability. However, INN-PAR reconstructs the high-frequency details of the ABP signal with much better accuracy. All these quantitative and qualitative results show the effectiveness of INN-PAR.

\subsection{Ablation Study}
To evaluate the effectiveness of our proposed method, we perform the following ablation experiments. \textbf{AE1:} Instead of joint learning the mapping between the signal and its gradient, we only learn the mapping between signals. Since the INN framework needs multi-channel input to maintain the invertibility, we repeat the single-channel PPG signal twice in the input. \textbf{AE2:} In the affine coupling layer, we extract the features at a single scale instead of multi-scale feature extraction. For this, we remove the downscaling and upscaling layers in INN-PAR and keep only the convolution and ReLU activation layers. \textbf{AE3:} In the loss function in Eqn. 4, we set the value of $\alpha$ to $0$. \textbf{AE4:} Our proposed method. For ablation experiments, we train on the Sensors dataset, and test on the BCG dataset. Table \ref{tab:ablation} shows the performance for different ablation experiments. As shown from the results of \textbf{AE1}, not using the gradient channel of PPG and ABP in INN-PAR severely degrades the performance. Also, the multi-scale feature extraction in MSCM is highly effective, as evident from the results of \textbf{AE2}. Moreover, constraining the gradient of the reconstructed ABP signal to that of the ground truth ABP signal is also effective, as demonstrated by the results of \textbf{AE3}. These results justify the design of INN-PAR as well as the loss function.

\begin{table}[hbt]
\fontsize{9.5}{11.5}\selectfont
\centering
\caption{Comparison results of different ablation experiments. The best values are highlighted. $\downarrow$ means a low value is desired.  SBP and DBP values are in mmHg unit.}
\begin{tabular}{ccccc}
\toprule
\multirow{2}{*}{Experiment}&
\multicolumn{2}{c}{Waveform}&
\multicolumn{1}{c}{SBP}& \multicolumn{1}{c}{DBP}\\
\cmidrule(lr){2-3}
\cmidrule(lr){4-4}
\cmidrule(lr){5-5}
&\multicolumn{1}{c}{MAE$\downarrow$}&\multicolumn{1}{c}{NRMSE$\downarrow$}& \multicolumn{1}{c}{MAE$\downarrow$}& \multicolumn{1}{c}{MAE$\downarrow$}\\
\midrule
\text{AE1} & 0.657 & 1.203 & 129.83 & 63.74 \\
\text{AE2} & 0.078 & 0.724 & 13.20 & 8.47 \\
\text{AE3} & 0.077 & 0.731 & 12.02 & 8.01 \\
\text{AE4} & \textbf{0.075}  & \textbf{0.701} & \textbf{11.96}  & \textbf{7.93}\\
\bottomrule
\end{tabular}
\label{tab:ablation}
\end{table}

\section{Conclusion}
\label{sec:conclusion}

This work introduces INN-PAR, an invertible network designed for the PAR task. Unlike other deep learning models, INN-PAR jointly learns the mapping between PPG and its gradient with those of the ABP signal, simultaneously capturing both forward and inverse mappings, thus, preventing information loss. Also, we propose MSCM to capture features at multiple scales. Our ablation experiments justify the design of INN-PAR. Experimental results demonstrate that INN-PAR outperforms the SOTA methods. Future research could investigate the application of INN-PAR to other physiological signal reconstruction tasks.

\clearpage
\bibliographystyle{IEEEtran}
\bibliography{ref}

\end{document}